\documentclass{article}
\usepackage{ijcai20}

\usepackage{times}
\usepackage{soul}
\usepackage{url}
\usepackage[hidelinks]{hyperref}
\usepackage[utf8]{inputenc}
\usepackage[small]{caption}
\usepackage{graphicx}
\usepackage{amsmath}
\usepackage{subfigure}
\usepackage{float}
\usepackage{bm}
\usepackage{amsmath,amssymb,amsfonts,amsthm}
\usepackage{booktabs}
\usepackage{algorithm}
\usepackage{algorithmic}

\title{Soft Hindsight Experience Replay}

\begin{document}
	
\maketitle
	
\begin{abstract}
    Efficient learning in the environment with sparse rewards is one of the most important challenges in Deep Reinforcement Learning (DRL). In continuous DRL environments such as robotic arms control, Hindsight Experience Replay (HER) has been shown an effective solution. However, due to the brittleness of deterministic methods, HER and its variants typically suffer from a major challenge for stability and convergence, which significantly affects the final performance. This challenge severely limits the applicability of such methods to complex real-world domains. To tackle this challenge, in this paper, we propose Soft Hindsight Experience Replay (SHER), a novel approach based on HER and Maximum Entropy Reinforcement Learning (MERL), combining the failed experiences reuse and maximum entropy probabilistic inference model. We evaluate SHER on Open AI Robotic manipulation tasks with sparse rewards. Experimental results show that, in contrast to HER and its variants, our proposed SHER achieves state-of-the-art performance, especially in the difficult HandManipulation tasks. Furthermore, our SHER method is more stable, achieving very similar performance across different random seeds.     
\end{abstract}
	
\section{Introduction}
	Reinforcement Learning (RL) combined with Deep Learning \cite{b1} has been shown an effective framework in a wide range of domains, such as playing video games \cite{b2}, defeating the best human player at the game of Go \cite{b3}, beating the professional teams of Dota2 \cite{OpenAI_dota}, Starcraft2 \cite{arulkumaran2019alphastar} and Quake3 \cite{jaderberg2019human}, as well as different robotic tasks \cite{levine2018learning,kalashnikov2018qt,andrychowicz2018learning}.\\
	\\
	However, many great challenges still exist in Deep Reinforcement Learning (DRL), one of which is to make the agent learn efficiently with sparse rewards. To tackle this challenge, one of the key concept is goal, which is proposed  in the early stage as the supplementary RL objective of the state-action value function $Q$ \cite{kaelbling1993learning}. For modern DRL, Universal Value Function Approximator \cite{schaul2015universal} is proposed to sample goals from some special states, which extends the definition of value function $V$ by not just over states but also over goals. In the same year, \cite{lillicrap2015continuous} developed  the Deep Deterministic Policy Gradient (DDPG), which has great performance in continuous control tasks such as manipulation and locomotion. Combining the above two methods, Hindsight Experience Replay (HER) \cite{andrychowicz2017hindsight} was proposed to replace the desired goals of training trajectories with the achieved goals, which additionally leverage the rich repository of the failed experiences. Utilizing HER, the RL agent can learn to accomplish complex robot manipulation tasks in the Open AI Robotics environment \cite{plappert2018multi}, which is nearly impossible to be solved with single RL algorithm like DDPG.\\
	\\
	In recent research works, the key concept of maximum entropy is universal to encourage exploration during training. In principle, inspired by the animal behavior in nature, Maximum Entropy Reinforcement Learning (MERL) can equivalently be viewed as $probability$ $matching$ with Probabilistic Graphical Models (PGMs) \cite{koller2009probabilistic} and a distribution defined by exponentiating the reward. Judged from the result, the optimized MERL objective is a modification of the standard RL objective that further adds an entropy term. The additional entropy term causes the agents to have stochastic behavior and non-zero probability of sampling every action. In this view, MERL agents behave carefully like natural animals and have a more comprehensive perspective on the whole environmental evaluation. Furthermore, by introducing PGMs, researchers developed corresponding RL algorithms for MERL, including Soft Policy Gradients \cite{haarnoja2017reinforcement}, Soft-Q Learning \cite{schulman2017equivalence} and Soft Actor-Critic \cite{haarnoja2018soft}.\\
	%\footnote{Our code is available at https://github.com/fslight1996/SHER.}
	\\
	Compared to deterministic methods, which are used in Open AI HER as basic framework, the MERL agents are not so greedy about rewards and using composable policies to gradually approximate the optimal solution \cite{eysenbach2019if,haarnoja2018acquiring}. MERL brings about stable behavior and prevents the convergence of policies to the local optima. These properties are highly beneficial for exploration in the environment with sparse rewards, such as the robotic manipulation tasks. Since realistic scenarios are full of random noise and multi-modal tasks, greedy tendency may lead to more instability and similarity to supervised learning which causes the policy to converge to the local optima. Actually, even in the Open AI Robotics simulated environment, since different epochs have different goals, the agents trained with HER algorithm usually fall into severe fluctuations or form a circular behavior between different epochs especially in the HandManipulation tasks. This phenomenon is quite common even using more CPU cores or more epochs for exploration, which demonstrates the instability and local optimality of HER. It is convinced to actually limit the efficient learning in the environment with sparse rewards.\\
	\\
	In this paper, we propose "Soft Hindsight Experience Replay" (SHER) to improve the training process in RL with sparse rewards and also evaluate SHER on the representative Open AI Robotics environment. On the basis of probabilistic inference model and the assumptions in HER, we derive the optimized soft objective formula for SHER and propose the corresponding algorithm. By introducing MERL into the HER algorithm framework, our main purpose is to efficiently improve the stability and convergence of the original HER through replacing deterministic RL with MERL. While in result, we found that SHER can achieve better performance compared to HER and its variant CHER \cite{fang2019curriculum}. Furthermore, we infer that the improvement is due to the improved stability of SHER according to our data analysis. In addition, the temperature of MERL may be a key parameter for the performance of SHER.     	
	\section{Related Work}
	HER is the first remarkable algorithm to make the agent learn efficiently in continuous environment with sparse rewards. After that, a series of algorithms are proposed based on HER. \cite{ding2019goal} combined HER with imitation learning and incorporated demonstrations to drastically speed up the convergence of policy to reach any goal. \cite{held2018automatic} proposed using GAN to generate different difficulty levels of goals to pick up the appropriate level of goals which automatically producing a curriculum.
	\cite{zhao2018energy} discovered an interesting connection between energy of robotic arms' trajectories and the finish of manipulation tasks. \cite{zhao2019curiosity} also proposed a curiosity-driven prioritization framework to encourage the sampling of rare achieved goal states. Furthermore, it is worth mentioning that \cite{zhao2019maximum} used weighted entropy to make a regularized transformation of the replay buffer, which is radically different from our work based on probabilistic inference model. \cite{fang2018dher} expanded the fix goal environment and solved the dynamic robotic fetch tasks.
	\cite{fang2019curriculum}  adaptively selected the failed experiences for replay according  to the curriculum-guided proximity and diversity functions of states. \cite{DBLP:journals/corr/abs-1902-00528} 
	utilized multi-agent DDPG to introduce a competition between two agents for better exploration. These works concentrate on failed experiences reuse, or the improvement of the replay buffer, while our work concentrate on improving the whole stability and performance of the training framework.\\
	\\
	%From a long-term perspective, the idea of leveraging hindsight experiences will persistently make contributions on issues about RL with sparse rewards. For instance, Open AI \cite{rauber2017hindsight} proposed Hindsight Policy Gradients to extend hindsight behavior from off-policy replay buffer to on-policy computations of policy gradients and Deepmind \cite{harutyunyan2019hindsight} proposed Hindsight Credit Assignment to explicitly assign credit to past decisions. We believe that RL with sparse rewards has great application prospects for the real world. %\cite{DBLP:journals/corr/abs-1902-00528} enable the learning process in a self-imitated manner and can be trained with supervised learning.We believe that RL with sparse rewards has great application prospects for the real world.
	On the other hand, quite significant progress has been made for MERL, the probabilistic RL method. \cite{haarnoja2017reinforcement} alters the standard maximum reward reinforcement learning objective with an entropy maximization term and the original objective can be recovered using a temperature parameter. \cite{haarnoja2018soft} demonstrated  a better performance while slowing compositional ability and robustness of the maximum entropy locomotion and robot manipulation tasks. \cite{eysenbach2019if} connected MERL with game theory and showed the substance and equivalent form of it.
	The above remarkable improvements lead us to apply MERL to HER. 
	
	\section{Preliminary}

	\subsection{Markov Decision Process} We consider an interacting process between an agent and an environment, including a set of states $\mathcal{S}$, a set of actions $\mathcal{A}$, a policy $\pi$ maps a state to an action, $\pi:\mathcal{S}\rightarrow \mathcal{A}$. At each time step $t$, the agent observes a state $s_t$ from $\mathcal{S}$ and chooses an action $a_t$ from $\mathcal{A}$ following a policy $\pi(a_t|s_t)$ and receives a reward $r_t$ from the environment. If the next state $s_{t+1}$ is only determined by a distribution $p\left(\cdot | s_{t}, a_{t}\right)$, the transitions $(s_t,a_t,s_{t+1},r_t)$ in the environment follow Markov Principles and the interacting process is a Markov Decision Process (MDP). Modeled as MDP, the agent can generate a trajectory $\tau=\left\{\left(s_{0}, a_{0}\right), \cdots,\left(s_{T-1}, a_{T-1}\right)\right\}$ of length $T$. The task of reinforcement learning is to pursue a maximum accumulated reward $R(\tau)=\sum_{t=0}^{\infty} \gamma^{t} r_{t}$, where $\gamma$ is the discount factor. Only in MDP environments, the solution of RL task can be realistic with value-based methods using $V(s)$ and $Q(s,a)$ or policy-based methods using $\pi(a|s)$.
	\subsection{Universal Value Function Approximators} 
\begin{figure}[ht]
		\centering
		\includegraphics[scale=0.18]{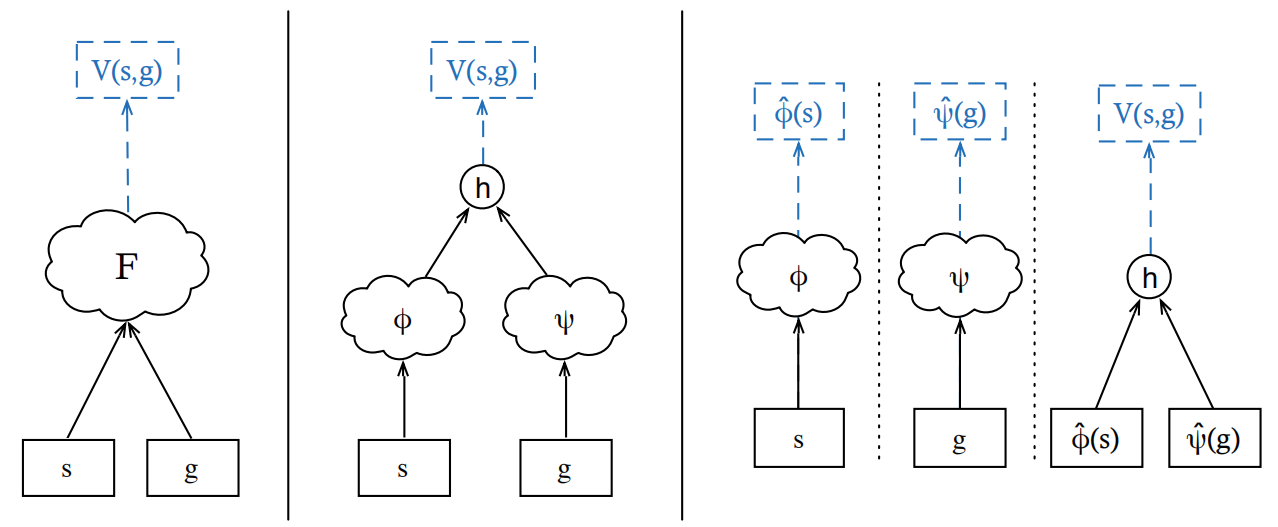}
		\caption{UVFA architectures}
		\label{Figure1}
	\end{figure}

\cite{schaul2015universal} proposed utilizing the concatenation of states $s\in\mathcal{S}$ and goals $g\in\mathcal{G}$ as higher dimensional universal states $(s,g)$ such that the value function approximators $V(s)$ and $Q(s,a)$ can be generalized as $V(s,g)$ and $Q(s,a,g)$. The goals can also be called goal states since in general $\mathcal{G}\subset\mathcal{S}$. Three architectures of UVFA are shown in Figure  \ref{Figure1}. We usually apply the first architecture in the algorithms and call RL in this framework Goal-conditioned RL or Multi-goal RL.    
	\\
	\subsection{Multi-goal RL and HER} Following UVFA method, in Multi-goal RL the optimized RL objective function $J$ can be expressed as follows:
	
	\begin{equation}\label{expect reward}
	\left.J(\theta)=\mathbb{E}_{s \sim \rho_{\pi}, a \sim \pi(a | s, g), g \sim G[} r_{g}(s, a, g)\right],
	\end{equation}
	where $\theta$ is the parameter to optimize $J$, $\rho_{\pi}$ is the normalized states distribution determined by the environment. According to the theorem of policy gradient, the gradient of $J(\theta)$ can be written as:
	
	\begin{equation}\label{multi-goal pg} 
	\nabla_{\theta} J(\theta)=\mathbb{E}_{\pi}\left[\nabla_{\theta} \log \pi(a | s, g) Q^{\pi}(s, a, g)\right],
	\end{equation}
	
	\begin{equation}\label{Q}
	Q^{\pi}(s, a, g)=\mathbb{E}_{\pi}\left[\sum_{t=1}^{\infty} \gamma^{t-1} r_{g}\left(s_{t}, a_{t}, g\right)\right].
	\end{equation}\\
	However, in the environments with sparse rewards, the Formula \ref{multi-goal pg} and Formula \ref{Q} are nearly impossible to be trained to convergence since the theorem of policy gradient depends on sufficient variability within the encountered rewards to accomplish the calculation of gradient ascent. In such environments, random exploration is unlikely to uncover this variability if goals are difficult to reach. To address this challenge, \cite{andrychowicz2017hindsight} proposed Hindsight Experience Replay (HER) including two key techniques, $reward\;shaping$ and $goal\;relabelling$. The key technique called $reward\;shaping$ is to make the reward function dependent on a goal $g \in G$, such that $r_{g}: S \times A \times G \rightarrow R$. In every episode a goal $g$ will be sampled and stay fixed for the whole episode. At every time step, as the MDP moves forward, $g$ is desired to be a target state guiding the RL agent to reach the state, so we apply the following function in the environment with sparse rewards:\\
	\begin{equation}\label{reward-shaping}
	r_{t}=r_{g}\left(s_{t}, a_{t}, g\right)=\left\{\begin{array}{c}{0, \text { if }\left|s_{t}-g\right|<\delta} \\ {-1, \text { otherwise }}\end{array}\right.
	\end{equation}
	where we can figure that this trick brings much more virtual rewards to support the training of the RL objective function. The reason for the technique to be effective is that the reward function Formula \ref{reward-shaping} based on Euclidean distance is strongly related to the possibility of final success although it is not shown through the real reward function. \\
	\\
	The other technique called $goal$ $relabelling$ is to replay each episode with different goals while not the one that the agent was trying to achieve. In HER paper, four schemes are proposed including $final$ --- replay the ones corresponding to the final state, $future$ --- replay with $k$ random states which come from the same episode as the transition being played, $episode$ --- replay with $k$ random states coming from the same episode as the transition being played, $random$ --- replay with $k$ random states encountered so far in the whole training procedure. In practice, we prefer to choose $future$ as the goal replay scheme.
	With $goal\;relabelling$, HER combine the failed experiences of separate episodes to generate new transitions with high dimensional concatenated states and take great advantage of the inner relationship between various failures.

	\section{Methodology}
	
	\subsection{Probabilistic Inference Model}
	
\begin{figure}[ht]
		\centering
		\includegraphics[scale=0.25]{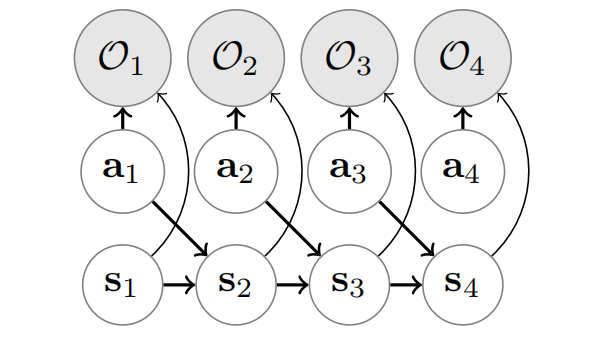}
		\caption{Probabilistic graphical model with optimality variables}
		\label{Figure2}
	\end{figure}

	The Figure \ref{Figure2} illustrates the principle of probabilistic inference \cite{koller2009probabilistic}, which is a widely recognized perspective on Maximum Entropy Reinforcement Learning (MERL). In the environment with sparse rewards, the explored proportion of the whole goal space is at a low level. Intuitively, we expect the introducing of probabilistic inference could improve the efficiency of exploring the uncertain goal space.
	\\
	In this model, we condition on the optimality variable being true, and then infer the most probable action sequence or the most probable action distributions. The additional variable is a binary random variable dependent both on states and actions, we choose the
	the distribution over this variable to be given by the following equation:
	
	\begin{equation}
	p\left(\mathcal{O}_{t}=1 | \mathbf{s}_{t}, \mathbf{a}_{t},\mathbf{g}\right)=\exp \left(r\left(\mathbf{s}_{t}, \mathbf{a}_{t},\mathbf{g}\right)\right),
	\end{equation}
	which is called $probability$ $matching$. As demonstrated in the Preliminary, the task in Multi-goal RL can be defined by the new reward function $r\left(\mathbf{s}_{t}, \mathbf{a}_{t}, \mathbf{g}_{\tau}\right)$ and solving a task typically involves recovering a policy $p\left(\mathbf{s}_{t}, \mathbf{a}_{t}, \mathbf{g}_{\tau} | \theta\right)$, which specifies a distribution conditioned on the parameterized Multi-goal state $\left(\mathbf{s}_{t}, \mathbf{g}_{\tau} | \theta\right)$. 
	The standard Multi-goal RL optimization task can be described by the following maximization:
	\begin{equation}\label{theta}
	\theta^{\star}=\arg \max _{\theta} \sum_{t=1}^{T} E_{\left(\mathbf{s}_{t}, \mathbf{a}_{t}, \mathbf{g}_{\tau}\right) \sim p\left(\mathbf{s}_{t}, \mathbf{a}_{t}, \mathbf{g}_{\tau} | \theta\right)}\left[r\left(\mathbf{s}_{t}, \mathbf{a}_{t}, \mathbf{g}_{\tau}\right)\right].
	\end{equation}
	The trajectory distribution of Multi-goal policy $p(\tau,\mathbf{g}_{\tau})$ can be expressed as:
	\begin{equation}\label{ptau}
	\begin{aligned}
	p(\tau,\mathbf{g}_{\tau})&=p\left(\mathbf{s}_{1}, \mathbf{a}_{t}, \ldots, \mathbf{s}_{T}, \mathbf{a}_{T}, \mathbf{g}_{\tau} | \theta\right)\\
	&=p\left(\mathbf{s}_{1}, \mathbf{g}_{\tau}\right) \prod_{t=1}^{T} p\left(\mathbf{a}_{t} | \mathbf{s}_{t}, \mathbf{g}_{\tau}, \theta\right) p\left(\mathbf{s}_{t+1} | \mathbf{s}_{t}, \mathbf{a}_{t}, \mathbf{g}_{\tau}\right).
	\end{aligned}
	\end{equation}
	Utilizing the optimal valuable $\mathcal{O}_{t:T}$, since Multi-goal RL requires the behavior policy to sample a fix goal for a single trajectory, we can derive the following formula:
	
	\begin{equation}\label{expreward}
	p\left(\tau, \mathbf{g}_{\tau},\mathbf{o}_{t:T}\right)=\mathcal{P}\cdot
	\exp \left(\sum_{t}^{T} r\left(\mathbf{s}_{t}, \mathbf{a}_{t},\mathbf{g}_{\tau}\right)\right),
	\end{equation}
	where $\mathcal{P}$ is the distributional probability of the inferred part of trajectory:
	
	\begin{equation}
	\mathcal{P}=\left[p\left(\mathbf{s}_{1},\mathbf{g}_{\tau}\right) \prod_{t}^{T} p\left(\mathbf{s}_{t+1} | \mathbf{s}_{t}, \mathbf{a}_{t}, \mathbf{g}_{\tau}\right)\right].
	\end{equation}
	
	\subsection{Soft Multi-goal RL} 
	
	Using Formula \ref{expreward}, after derivation quite similar to which in \cite{haarnoja2017reinforcement}, we derived the Multi-goal soft Bellman backup:
	
	\begin{equation}
	\resizebox{.9\hsize}{!}{$Q\left(\mathbf{s}_{t}, \mathbf{a}_{t}, \mathbf{g}\right) \leftarrow r\left(\mathbf{s}_{t}, \mathbf{a}_{t},\mathbf{g}\right)+\gamma E_{\mathbf{s}_{t+1} \sim p\left(\mathbf{s}_{t+1} | \mathbf{s}_{t}, \mathbf{a}_{t}, \mathbf{g}\right)}\left[V\left(\mathbf{s}_{t+1}, \mathbf{g}\right)\right]$},
	\end{equation}\\
	where the goal of $Q\left(\mathbf{s}_{t}, \mathbf{a}_{t}, \mathbf{g}\right)$ and the goal of $V\left(\mathbf{s}_{t+1}, \mathbf{g}\right)$ are the same owing to the reason that HER synchronously updates the goal of $s_{t}$ and $s_{t+1}$ in the same transition, which is the cornerstone of our theorem. Otherwise, the formula cannot be established and there will be plenty of extra computation for transitions of different goals. \\
	\\
	In order to minimize the gap between explored goal distribution $p(\tau, g_{\tau})$ and optimal goal distribution $\hat{p}(\tau,g_{\tau})$, we apply KL divergence as the following objective:
	
	\begin{equation}
	\resizebox{1.\hsize}{!}{$-D_{\mathrm{KL}}(\hat{p}(\tau,g_{\tau}) \| p(\tau,g_{\tau}))=E_{\tau \sim \hat{p}(\tau,g_{\tau})}[\log p(\tau,g_{\tau})-\log \hat{p}(\tau,g_{\tau})$},
	\end{equation}
	and the result is given by:
	\begin{equation}
	\sum_{t=1}^{T} E_{\left.\left(\mathbf{s}_{t}, \mathbf{a}_{t},g\right) \sim \hat{p}\left(\mathbf{s}_{t}, \mathbf{a}_{t},g\right)\right)}\left[r\left(\mathbf{s}_{t}, \mathbf{a}_{t},g\right)+\mathcal{H}\left(\pi\left(\mathbf{a}_{t} | \mathbf{s}_{t},g\right)\right)\right].
	\end{equation}
	The optimal inferred policy is given by:
	\begin{equation}\label{oppi}
	\pi\left(\mathbf{a}_{T} | \mathbf{s}_{T},g\right)=\exp \left(r\left(\mathbf{s}_{T}, \mathbf{a}_{T},g\right)-V\left(\mathbf{s}_{T},g\right)\right).
	\end{equation}
	To optimize the policy in Formula \ref{oppi}, we have to optimize the $V\left(\mathbf{s}_{T},g\right)$ first and the soft Multi-goal RL value function can be optimized by:
	\begin{equation}
	\resizebox{.95\hsize}{!}{$J_{V}(\psi)=\mathbb{E}_{\left(\mathbf{s}_{t},g\right) \sim \mathcal{D}}\left[\frac{1}{2}\left(V_{\psi}\left(\mathbf{s}_{t},g\right)-\mathbb{E}_{\mathbf{a}_{t} \sim \pi_{\phi}}\left[Q_{\theta}\left(\mathbf{s}_{t}, \mathbf{a}_{t},g\right)-\log \pi_{\phi}\left(\mathbf{a}_{t} | \mathbf{s}_{t},g\right)\right]\right)^{2}\right]$}
	\end{equation}
	where $\mathcal{D}$ represents the replay buffer to store the transitions $\left(s_{t}\left\|g, a_{t}, r_{t}, s_{t+1}\right\| g\right)$. With the above formulas, we obtain the final two objectives of optimization --- Multi-goal V-loss gradient and $\pi$-loss gradient given by:
	\begin{equation}
	\resizebox{.95\hsize}{!}{$\hat{\nabla}_{\psi} J_{V}(\psi)=\nabla_{\psi} V_{\psi}\left(\mathbf{s}_{t},g\right)\left(V_{\psi}\left(\mathbf{s}_{t},g\right)-Q_{\theta}\left(\mathbf{s}_{t}, \mathbf{a}_{t},g\right)+\log \pi_{\phi}\left(\mathbf{a}_{t} | \mathbf{s}_{t},g\right)\right)$},
	\end{equation}
	\begin{equation}
	\resizebox{.95\hsize}{!}{$\begin{aligned} \hat{\nabla}_{\phi} J_{\pi}(\phi) &=\nabla_{\phi} \log \pi_{\phi}\left(\mathbf{a}_{t} | \mathbf{s}_{t},g\right) \\ &+\left(\nabla_{\mathbf{a}_{t}} \log \pi_{\phi}\left(\mathbf{a}_{t} | \mathbf{s}_{t},g\right)-\nabla_{\mathbf{a}_{t}} Q\left(\mathbf{s}_{t}, \mathbf{a}_{t},g\right)\right) \nabla_{\phi} f_{\phi}\left(\epsilon_{t} ; \mathbf{s}_{t},g\right) \end{aligned}$},
	\end{equation}
	where $\epsilon_{t}$ is an input noise vector and $ f_{\phi}\left(\epsilon_{t} ; \mathbf{s}_{t},g\right)$ is a neural network to reparameterize the Multi-goal policy.\\ 
	\\
	Hence, we propose the algorithm of Soft Multi-goal RL as follows:
	\begin{algorithm}[H]
		\caption{Soft Hindsight Experience Replay}
		\label{alg1}
		\begin{algorithmic}[1]
			\STATE Input: initial policy parameters $\theta$, 
			Q-function parameters $\phi_1$, $\phi_2$,
			V-function parameters $\psi$, 
			empty replay buffer $\mathcal{R}$,
			a strategy $\mathcal{S}$ for sampling goals for replay \\
			available:any optimization algorithm $\mathcal{A}$ on replay buffer e.g. PER, EBP, CDP, MEP, CHER
			\STATE  Initialize replay buffer $\mathcal{R}$,
			Set target parameters equal to main parameters $\psi_{\text{targ}} \leftarrow \psi$
			\FOR{episode = 1,M} 
			\STATE Sample a goal $g$, initial state $s_{0}$
			\FOR {$t$ = 0, $T-1$} 
			\STATE Sample an action from \\$a_{t} \sim \underset{\pi_{\theta}}{\mathrm{E}}\left[Q^{\pi_{\theta}}(s_{t},\cdot,g )-\alpha \log \pi_{\theta}(\cdot|s_{t},g)\right]$
			\STATE Execute $a_{t}$ in the environment and get next state $s_{t+1}$
			\ENDFOR
			\FOR {$t$ = 0, $T-1$}
			\STATE $r_{t}:=r\left(s_{t}, a_{t}, g\right)$
			\STATE Store the transition $\left(s_{t}\left\|g, a_{t}, r_{t}, s_{t+1}\right\| g\right)$ in replay buffer $\mathcal{R}$
			\STATE Sample a set of additional goals for replay $G:=S$
			\FOR {$g^{\prime} \in G$} 
			\STATE $r^{\prime}:=r\left(s_{t}, a_{t}, g^{\prime}\right)$
			\STATE Store the transition $\left(s_{t}\left\|g^{\prime}, a_{t}, r^{\prime}, s_{t+1}\right\| g^{\prime}\right)$ in $\mathcal{R}$
			\STATE Using $\mathcal{A}$ to adjust the weight and priority of transitions, trajectories or achieved goals
			\ENDFOR
			\ENDFOR
			\FOR{$t = 1,N$}
			\STATE Sample a minibatch B from the replay buffer $\mathcal{R}$
			\FOR{each transition in B}
			\STATE Multi-goal Q-targets and V-targets updating 
			\begin{small}
				
				$y_q (r,s',g) = r(s,g) + \gamma (1-d) V_{\psi_{\text{targ}}}(s',g)$ \\
				$y_v (s,g) = \min_{i=1,2} Q_{\phi_i} (s, \tilde{a},g) - \alpha \log \pi_{\theta}(\tilde{a}|s,g)$
				
			\end{small}	
			
			\STATE Multi-goal V-loss and $\pi$-loss gradient descent

			$\nabla_{\psi} \frac{1}{|B|}\sum_{s \in B} \left( V_{\psi}(s,g) - y_v(s,g) \right)^2$
			\\
			
			\resizebox{.8\hsize}{!}{$\nabla_{\theta} \frac{1}{|B|}\sum_{s \in B} \Big( Q_{\phi,1}(s,g,\tilde{a}_{\theta}(s))-\alpha \log \pi_{\theta} \left(\left. \tilde{a}_{\theta}(s,g) \right| s,g\right) \Big)$}
			\STATE Target value network updating \\
			$\psi_{\text{targ}}\leftarrow \rho \psi_{\text{targ}} + (1-\rho) \psi$
			\ENDFOR
			\ENDFOR
			\ENDFOR
		\end{algorithmic}
		
	\end{algorithm}

	\section{Experiments}

	\subsection{Environments}
\begin{figure*}[ht]
		\centering
		\subfigure[FetchReach]{
			\includegraphics[width=0.23\textwidth]{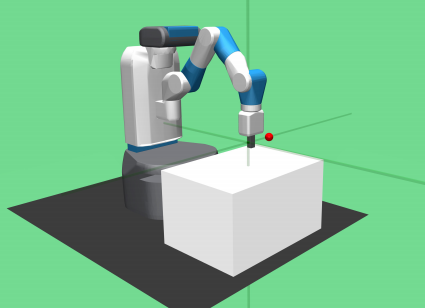} 
		}
		\subfigure[FetchPush]{
			\includegraphics[width=0.23\textwidth]{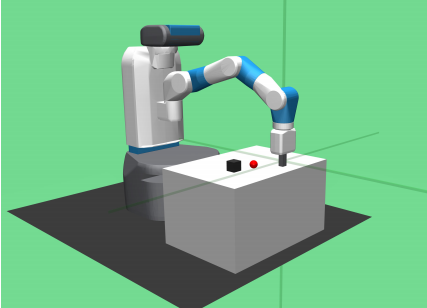}
		}
		\subfigure[FetchSlide]{
			\includegraphics[width=0.23\textwidth]{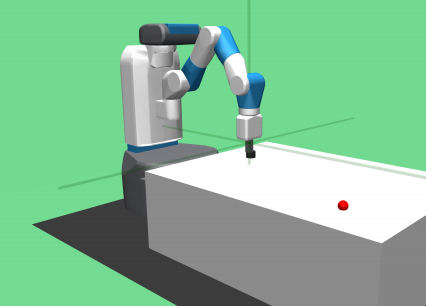}
		}
		\subfigure[FetchPickAndPlace]{
			\includegraphics[width=0.23\textwidth]{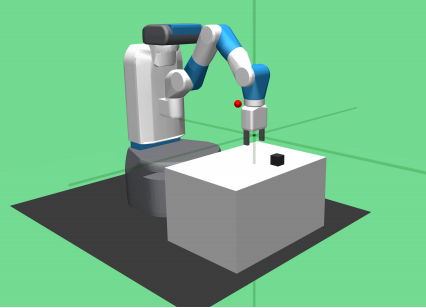}
		}
		\subfigure[HandReach]{
			\includegraphics[width=0.23\textwidth]{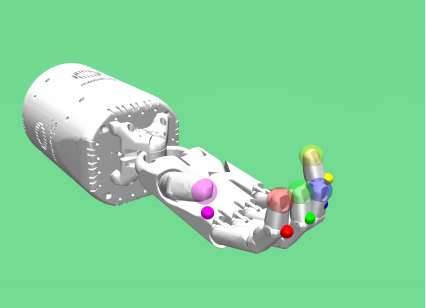}
		}
		\subfigure[HandManipulateBlock]{
			\includegraphics[width=0.23\textwidth]{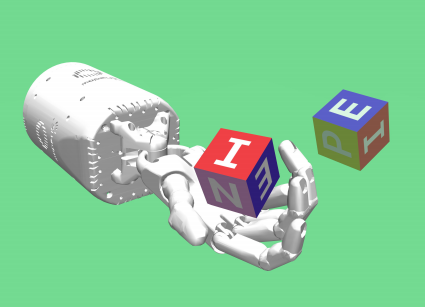}
		}
		\subfigure[HandManipulateEgg]{
			\includegraphics[width=0.23\textwidth]{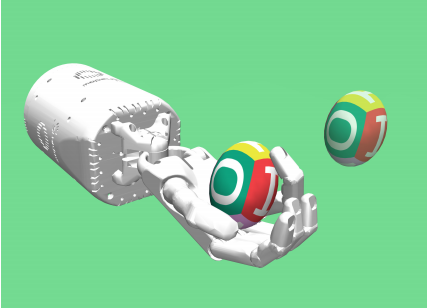}
		}
		\subfigure[HandManipulatePen]{
			\includegraphics[width=0.23\textwidth]{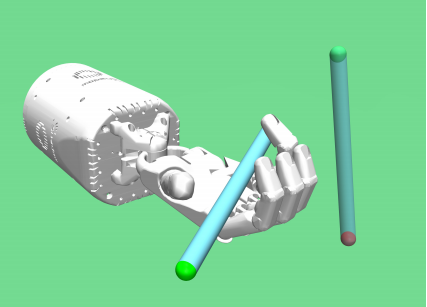}
		}

		\caption{the Open AI Robotics environment for RL with sparse rewards}
		\label{robotics}
	\end{figure*}

	We evaluate SHER and compare to the baselines on several challenging robotic manipulation tasks in simulated Mujoco environments Robotics \cite{plappert2018multi} as the Figure \ref{robotics} shows, including two kinds of tasks, Fetch robotic arm tasks and Shadow Dexterous Hand tasks. Both two kinds of tasks have sparse binary rewards and follow a Multi-goal RL framework in which the agent is told what to do using an additional input. The agent obtains a reward of 0 if the goal has been
	achieved and -1 otherwise.\\
	\\
	\textbf{FetchEnv} The Fetch environments are based on the 7-DoF Fetch robotic arm, which has a two-fingered parallel gripper. In all Fetch tasks, the goals are 3-dimensional vectors describing the desired positions of the object and actions are 4-dimensional vectors including 3 dimensions to specify the desired gripper movement and the last dimension to control opening and closing of the gripper. Observations include the Cartesian position of the gripper, its
	linear velocity as well as the position and linear velocity of the robot’s gripper. The task of FetchEnv is to move, slide or place something to the desired position.\\
	\\
	%\textbf{FetchReach} To move the gripper to a target position.\\
	%\textbf{FetchPush} To move the box to a target location and robot fingers are locked.\\
	%\textbf{FetchSlide} To hit the puck to slide it to the target position which is outside of the robot's reach.\\
	%\textbf{FetchPickAndPlace} To grasp a box and move it to the target location which may be located in the air.\\
	\\
	\textbf{HandEnv}  The Shadow Dexterous Hand is an anthropomorphic robotic
	hand with 24 DoF of which 20 joints can be controlled independently and the remaining ones are coupled joints. The actions are 20-dimensional vectors containing the absolute position control for all non-coupled joints of the hand. Observations include the 24 positions and velocities of the
	robot’s joints. The task of HandEnv is to manipulate something only by fingers to the desired position and angle. It can be seen from the description that the HandEnv are much more difficult than FetchEnv and actually amongst $the$ $most$ $difficult$ ones of $all$ the Open AI Gym environments.\\
	\\
	We run the experiments with three algorithms: \\
	\\
	\textbf{HER} The original framework of Multi-goal RL. \\
	\textbf{CHER} The baseline with the best performance among the related work utilizing curriculum learning trick\cite{fang2019curriculum}.\\
	\textbf{SHER} Our work utilizing Soft Multi-goal RL, without any trick.\\
	%\textbf{HandReach} To make the mean distance between fingertips and their desired position less than 1 cm.
	
	%\textbf{HandManipulateBlock} To manipulate the block such that the distance between the block’s position and its desired position is
	%less than 1 cm and the difference in rotation is less than 0.1 rad.
	
	%\textbf{HandManipulateEgg} To manipulate the egg such that the distance between the block’s position and its desired position is less than 1 cm and the difference in rotation is less than 0.1 rad.
	
	%\textbf{HandManipulatePen} To manipulate the Pen such that the distance between the block’s position and its desired position is less than 5 cm and the difference in rotation is less than 0.1 rad.\\

	\begin{figure*}[h]
		\centering
		\subfigure[FetchReach]{
			\includegraphics[width=0.23\textwidth]{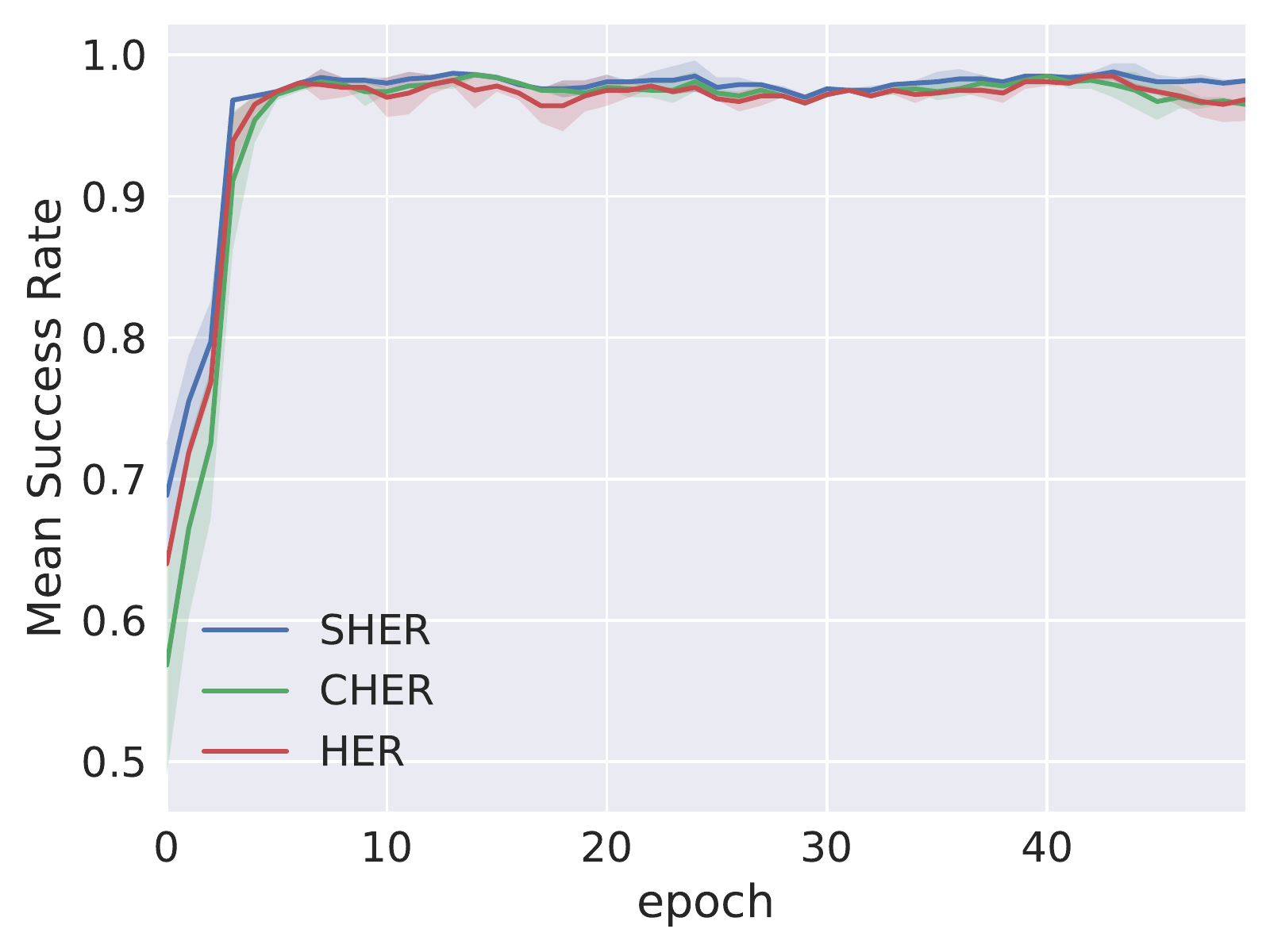} 
		}
		\subfigure[FetchPush]{
			\includegraphics[width=0.23\textwidth]{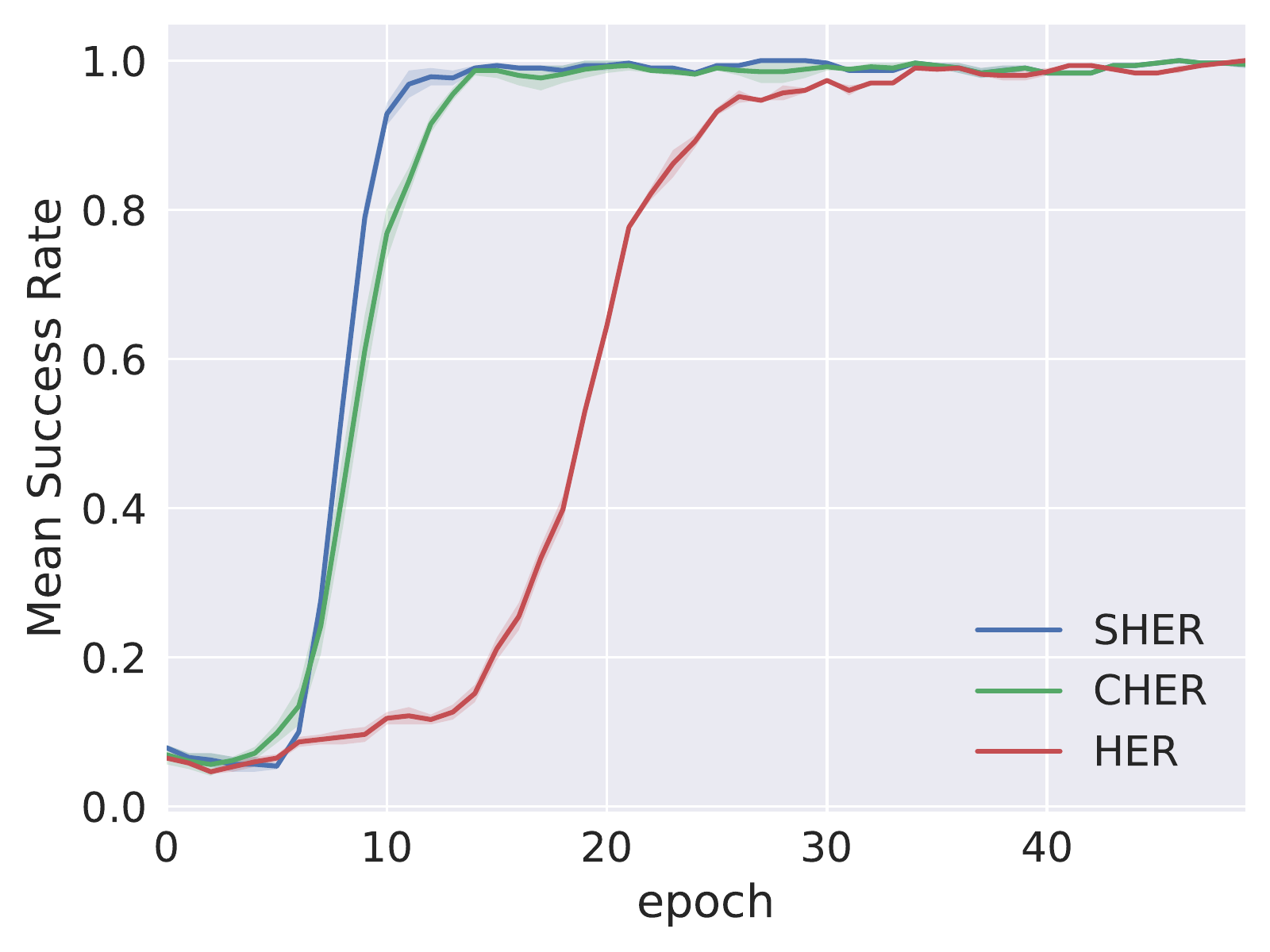}
		}
		\subfigure[FetchSlide]{
			\includegraphics[width=0.23\textwidth]{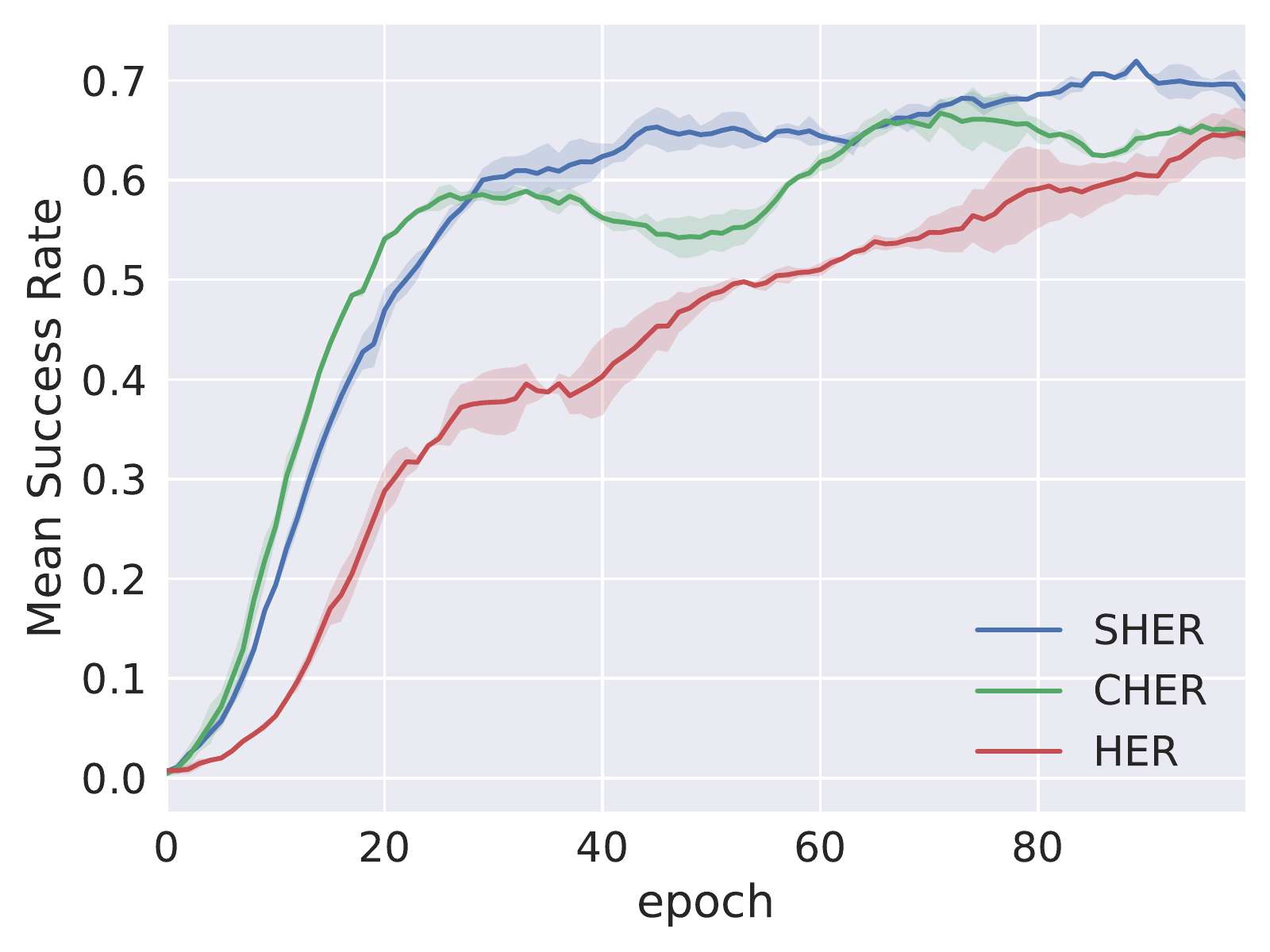}
		}
		\subfigure[FetchPickAndPlace]{
			\includegraphics[width=0.23\textwidth]{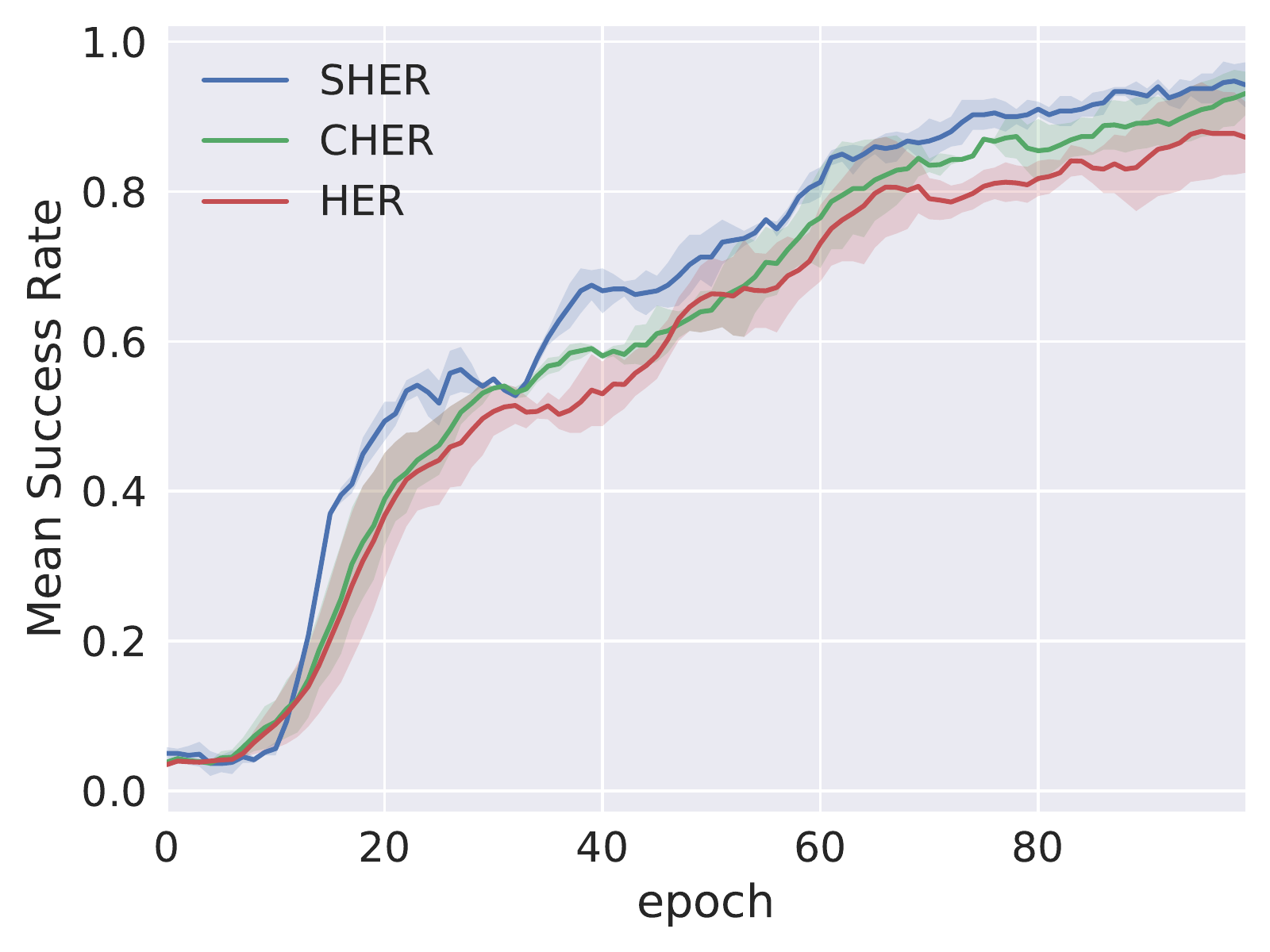}
		}
		\subfigure[HandReach]{
			\includegraphics[width=0.23\textwidth]{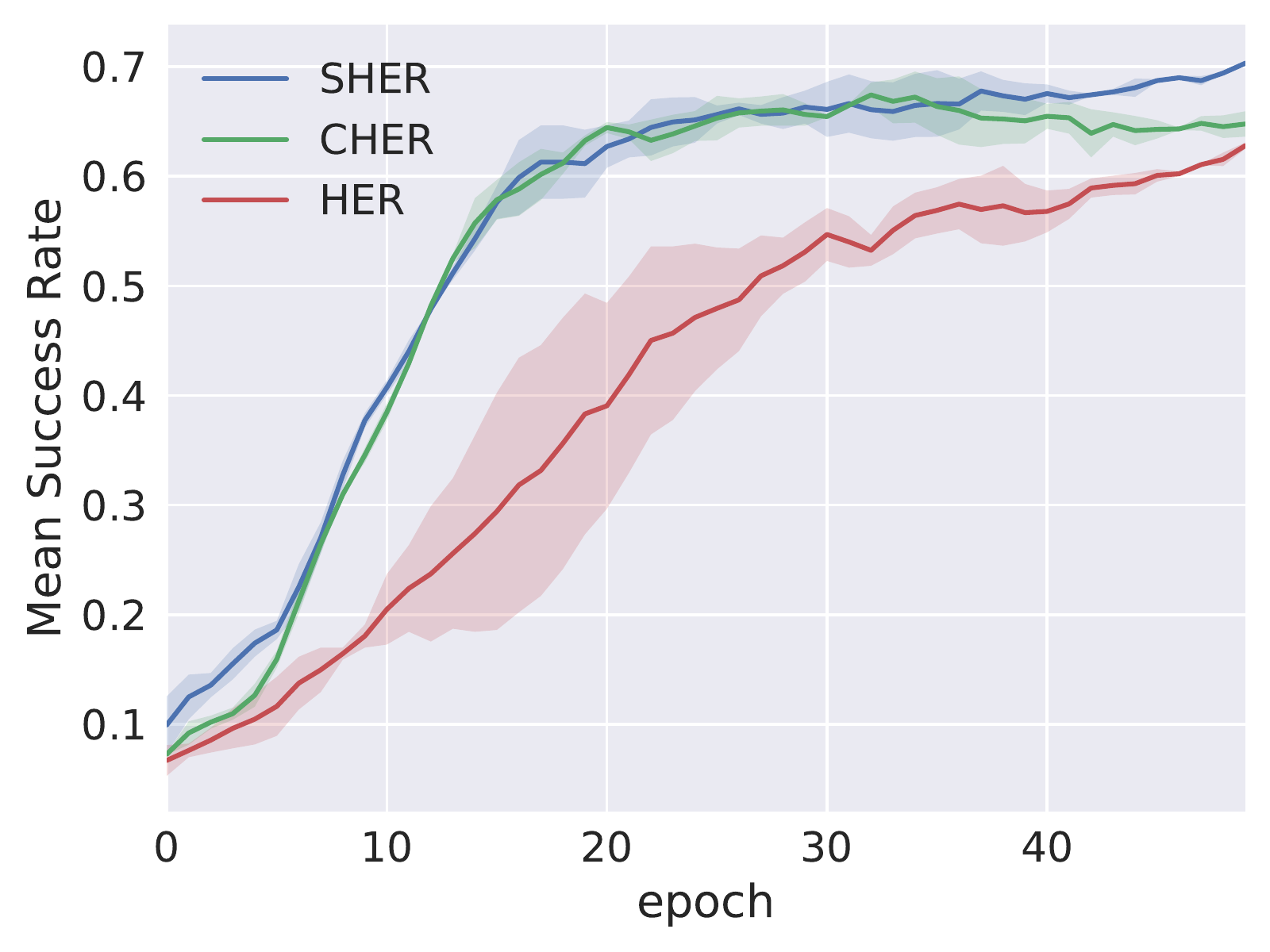}
		}
		\subfigure[HandManipulateBlockRotate]{
			\includegraphics[width=0.23\textwidth]{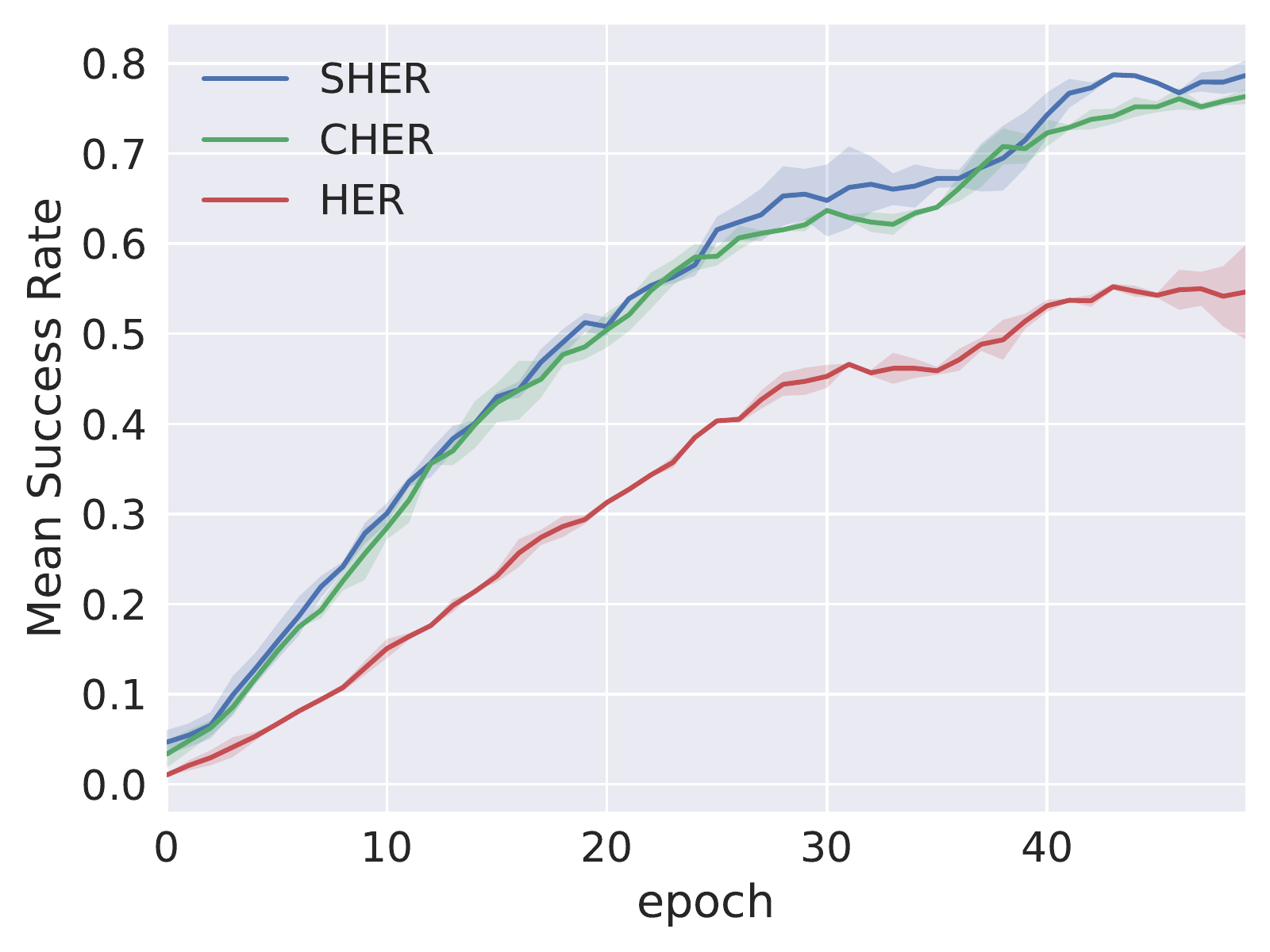}
		}
		\subfigure[HandManipulateEggFull]{
			\includegraphics[width=0.23\textwidth]{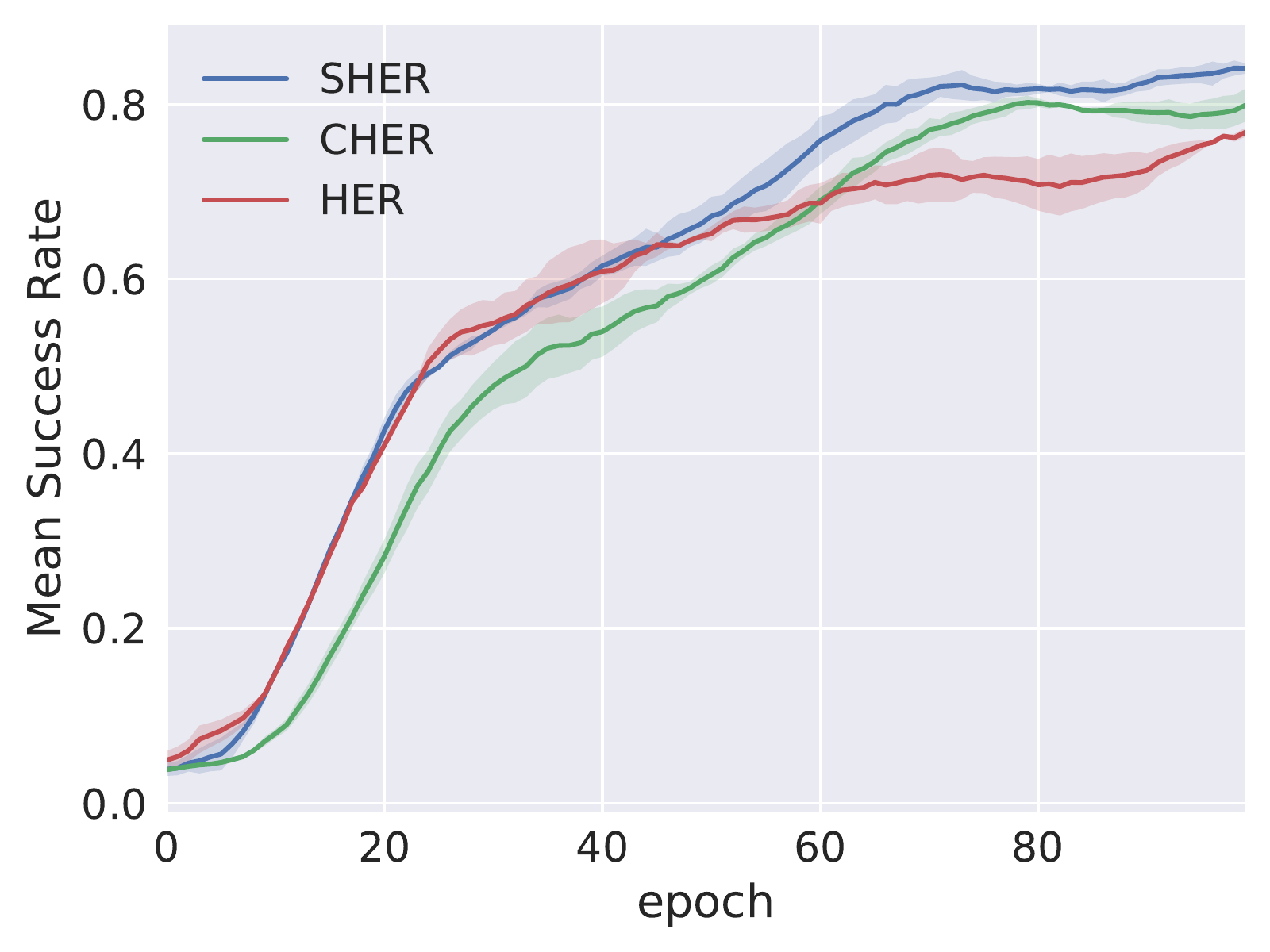}
		}
		\subfigure[HandManipulatePenRotate]{
			\includegraphics[width=0.23\textwidth]{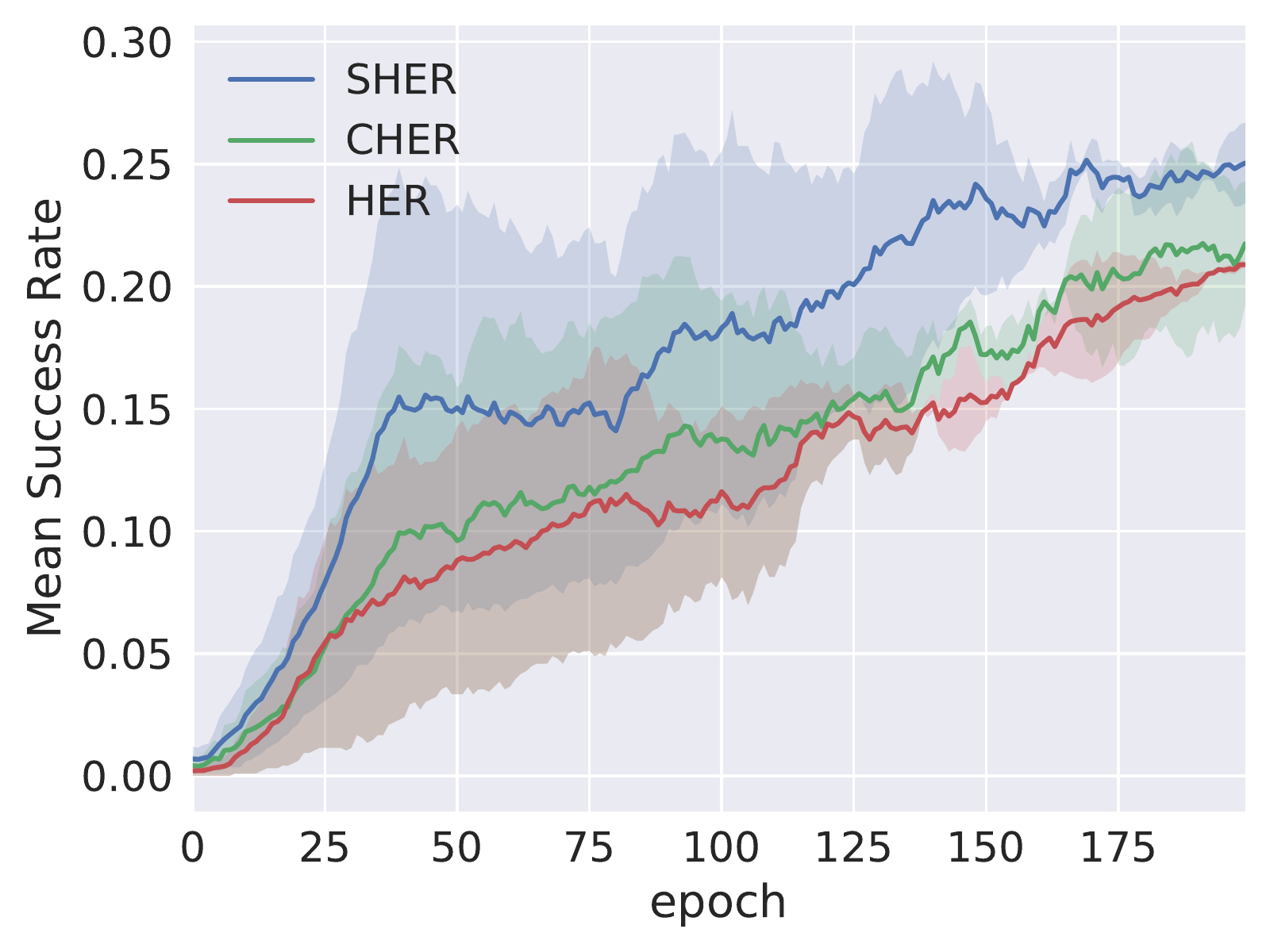}
		}

		\caption{Benchmark results for the Open AI Robotics environment}
		\label{bench}
		
	\end{figure*}
	
	\begin{table*}
		\centering
		\begin{tabular}{lllllllll}
			\hline
			Method & FReach & FPush & FPandP & FSlide & HReach & HEgg &HBlock &HPen  \\
			\hline
			HER    & 0.021   & 0.035  & 0.156 & 0.218&0.195&0.145&0.152& 0.074 \\
			CHER   & 0.018  & 0.022 & 0.141 &0.153&0.174&0.109&\textbf{0.081}& 0.066  \\
			SHER(Ours)   & \textbf{0.011}  &\textbf{0.014}  & \textbf{0.079} & \textbf{0.116}& \textbf{0.137}& \textbf{0.098}& 0.093& \textbf{0.059} \\
			
			\hline
		\end{tabular}
		\caption{Mean $\mathcal{\delta S}$ values in different environments : use FReach short for FetchReach environment and so on}
		\label{RG}
	\end{table*}
	
	\subsection{Benchmark Performance}
	In the benchmark experiment, the better mean success rate represents for better performance to accomplish robotic tasks.
	Now we compare the mean success rate in Figure \ref{bench}, where the shaded area represents the standard deviation since we use different random seeds.  The agent trained with SHER shows a better benchmark performance at the end of the training. It is surprising that SHER not only surpass HER but also can be better than CHER without any specialized trick.

	\subsection{Stability and Convergence}
	
	From Figure \ref{bench}, we can see that SHER converges faster in all eight tasks than both HER and CHER, which demonstrates the great convergence of SHER.
	Although we have achieved better performance than baselines in benchmark performance, for our main purpose, it is still not convenient to intuitively verify that SHER has better stability than HER and CHER. To show this in some way, we propose the formula

	\begin{equation}
	\mathcal{\delta S}=|S_{train}-S_{test}|
	\end{equation}
	to measure the stability of different goals for three algorithms, where $S_{train}$ and $S_{test}$ stand for the success rate of training set and testing set in the same episode that no more than 1. If the values of $\mathcal{\delta S}$ are smaller along the whole training process for an algorithm, we can infer in some way that the algorithm has better stability since goals are sampled with different random seeds in different epochs. From Table \ref{RG}, we can figure that in 7 of 8 environments the $\mathcal{\delta S}$ values of SHER are smaller than HER and CHER, which means SHER has better stability in the Open AI Robotics environment.

	\subsection{Temperature Parameter}
	
\begin{figure}
		\centering
		\includegraphics[scale=0.3]{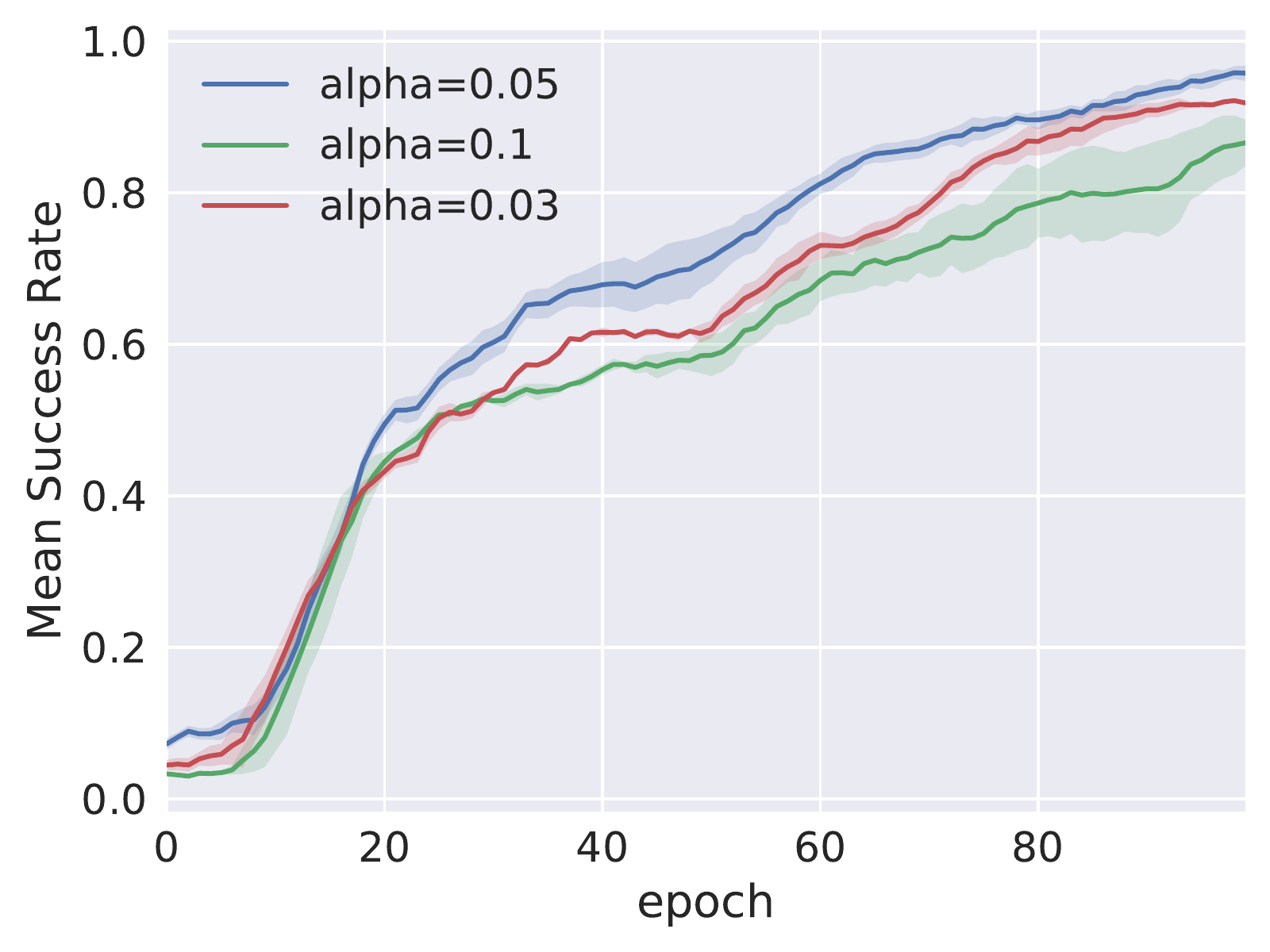}
		\caption{Experiments on different values of temperature}
		\label{temperature}
	\end{figure}

	We found that the temperature $\alpha$ which decides the ratio between deterministic RL and probabilistic RL, has a significant impact on the performance of SHER. According to our experiments, different values of $\alpha$ will be needed in different environments. If the value of $\alpha$ is not in a correct range, the agent usually can not be trained to the convergence. Figure \ref{temperature} is an example in FetchSlide:\\
	\\
	As Figure \ref{temperature} shows, three SHER agents are trained with $\alpha=0.03,0.05,0.1$. Finally, the agent with $\alpha=0.05$ has the best performance. The agent with 0.03 is not as good as 0.05 but better than  the agent with 0.1. However, in other environments  0.05 may not be the best value for $\alpha$.  It will be needed for further research on how to find a suitable $\alpha$.

	\section{Conclusion}
	
	The main contributions of this paper are summarized as follows: (1) We introduce "Soft Hindsight Experience Replay" as an adaptive combination of HER and MERL, which is the first work that prove the probabilistic inference model can be used in the environment with sparse rewards; (2) We show that SHER can exceed HER and CHER to achieve the state-of-the-art performance on Robotics without any trick; (3) We show that the training process of SHER are more stable and easier to converge to the optima due to the MERL framework; (4) Since the research of Multi-goal RL focus more on the improvement of experience replay, our work can be an excellent baseline or basic framework to accelerate and optimize the training process. 
	
	%(5) We are convinced that SHER can play an important role in realistic robotic control, which needs further research.

	\bibliographystyle{named}
	\bibliography{SHER}

\end{document}